\documentclass[12pt]{article}
\usepackage{fullpage}
\usepackage{graphicx}
\usepackage{amssymb}

\title{Memory Augmented Large Language Models are Computationally Universal}
\author{Dale Schuurmans\\
Google Brain \& University of Alberta}
\date{}

\begin{document}

\maketitle

\begin{abstract}

We show that transformer-based large language models
are computationally universal when augmented with an external memory.
Any deterministic language model that conditions
on strings of bounded length is equivalent to a finite automaton,
hence computationally limited.
However, augmenting such models with a read-write memory
creates the possibility of processing arbitrarily large inputs
and, potentially, simulating any algorithm.
We establish that an existing large language model, Flan-U-PaLM~540B,
can be combined with an associative read-write memory
to exactly simulate
the execution of
a universal Turing machine, $U_{15,2}$.
A key aspect of the finding is that it does not require
any modification of the language model weights.
Instead, the construction relies solely
on designing a form of stored instruction computer that 
can subsequently be programmed with a specific set of prompts.

\end{abstract}

\section{Introduction}
\label{sec:intro}

Interest in large language models has grown dramatically since the early
successes of GPT-2, GPT-3 and InstructGPT
\cite{radfordetal19,brownetal20,ouyangetal22},
and more recently with the popularity of ChatGPT \cite{schulmanetal22}.
Beyond simple question answering, where an input string posing a question
might elicit an output string containing a reasonable answer,
an important discovery has been the emergence of \emph{in-context learning},
where prepending a question with a set of related (question, answer) pairs
significantly improves question answering accuracy \cite{radfordetal19}.
Even adding a natural language instruction before
example pairs appears to further enhance language model capabilities
\cite{brownetal20}.
More recently, \emph{chain of thought prompting}
has been found to improve question answering ability
in scenarios where multiple
reasoning steps are required to arrive at a final answer,
such as answering math word problems \cite{weietal22b}.

Despite these results, current transformer-based large language
models remain fundamentally limited as they can only condition
on an input string of bounded length, such as 4096 tokens.
This makes such models formally equivalent to finite automata,
hence restricted in the computations they can express.
However, recent works have begun to investigate techniques for
chaining multiple calls to a language model
by processing model outputs then passing these back as subsequent inputs
to the model.
An example is \emph{least to most prompting},
where a complex reasoning question is answered first by prompting the model
to produce simpler sub-questions,
then passing each sub-question and resulting answer back into the model
to help answer subsequent sub-questions,
until a final answer is reached \cite{zhouetal22}.
Another example is work on \emph{language model cascades}
that investigates various strategies
for processing model outputs and feeding these
as inputs to subsequent language model
calls \cite{dohanetal22}.
Such works raise the question of whether
augmenting a language model with an external feedback loop
is merely useful, or fundamentally expands the range of computations
that can be performed.
To investigate this question, we consider
augmenting a language model with an external read-write memory
and ask whether
this confers the ability to simulate any algorithm on any input.

This paper gives an affirmative answer by establishing computational
universality for a specific large language model, Flan-U-PaLM~540B 
\cite{chungetal22},
augmented with an associative read-write memory.
A key aspect of the result is that it is achieved by developing
a simple form of \emph{stored instruction computer}
\cite{vonneumann45} that connects the language model to an associative memory,
then follows a simple instruction cycle where
the next input prompt to be passed to the language model
is retrieved from memory,
the output of the language model is parsed to recover any variable assignments
that are then stored in the associative memory,
and the next instruction is retrieved
(i.e., the next input prompt to be provided to the language model).
Each parsing step 
between the language model and memory
is performed by a simple regular expression match
(i.e., a finite automaton).

Once a stored instruction computer has been created,
a specific ``prompt program'' is designed to drive the system to simulate 
a universal Turing machine $U_{15,2}$ \cite{neary08,nearywoods09}.
Proving the fidelity of the simulation reduces to checking a
finite set of prompt-result behaviours and verifying that the language model
produces the correct output for each of the finite set of possible input
prompt strings it might encounter.
That is, although the overall input-output behaviour of the
Flan-U-PaLM~540B model
is not fully understood, a sufficiently reliable subset of its
input-output map can be isolated and controlled
to simulate a universal computer.
Importantly, this result does not involve any additional ``training'' of the
language model (i.e., no modification of its pre-trained weights),
but instead relies solely on providing specific
prompt strings to the model
and parsing its outputs to determine values to be saved in memory.

\section{Stored instruction computer}
\label{sec:computer}

As noted, there are many ways to orchestrate feedback between the
outputs of a language model and subsequent input prompts
\cite{zhouetal22,dohanetal22}.
In developing a simple feedback loop we would like to minimize
external processing and perform 
as much of the computation with the language model as possible,
while still supporting computational universality.
To achieve this, we consider a simple form of
\emph{stored instruction computer} \cite{vonneumann45},
where the language model plays the role of a central processing unit (CPU),
and the random access memory (RAM) is
supplied by an external associative memory.
Such an architecture allows for a simple interaction loop that can support
general computation and convenient programmability.
In this architecture,
the external associative memory is a simple ``dictionary'', {\tt MEMORY},
that maps unique keys to values,
or equivalently, maps variable names to values, or address locations to values.
Unlike the RAM in a physical computer, 
variable names will be strings
(i.e., finite length sequences of symbols from a finite alphabet)
to support convenient interaction with a language model,
while values will be strings or integers.

To ensure that computational universality does not follow simply from external
processing capability, 
all interaction between the language model and the memory 
will be restricted to
\emph{finite state computation},
such as simple regular expression parsers.

All code below will be given in Python 3
using the standard regular expression library {\tt re}.
Note that the regular expressions used for pre and post processing
are rudimentary and can easily be improved in several ways;
the versions given are merely sufficient to establish the main points
in this paper.

\subsection{Post processing language model outputs}

The output string from the language model will be
parsed by a simple regular expression that detects assignments in the form
{\tt variable\_name = "value"},
which are then applied to the associative memory as
{\tt MEMORY[variable\_name] = "value"}.
The variable assignment function {\tt assignments} is shown below.

\begin{verbatim}
def assignments(string):
  global MEMORY
  regex = '(?s)(?:((?:\w|\-)+)\s*=\s*(?:\"((?:.*\n)|(?:[^\"]*))\"))(.*)'
  matches = re.findall(regex, string)
  suffix = ''
  while len(matches) > 0:
    label, value, suffix = matches[0]
    MEMORY[label] = value
    matches = re.findall(regex, suffix)
  return suffix
\end{verbatim}

Additionally, splicing is allowed in value strings before being assigned to
memory;
that is, we include a regular expression parser that detects occurrences
of the pattern {\tt \%[variable\_name]} in any value string, 
replacing any such occurrence with the string at memory location
{\tt variable\_name}, i.e., {\tt MEMORY[variable\_name]},
before assignment.
The substitution function {\tt substitute} is shown below.

\newpage

\begin{verbatim}
def substitute(string, char):
  global MEMORY, BLANK
  regex = f"(?s)(.*?)(?:{char}\[(\w+)\])(.*)"
  matches = re.findall(regex, string)
  string = suffix = ''
  while len(matches) > 0:
    prefix, label, suffix = matches[0]
    if label not in MEMORY:
      MEMORY[label] = BLANK # new label has BLANK value by default
    string += prefix + str(MEMORY[label])
    matches = re.findall(regex, suffix)
  string += suffix
  return string
\end{verbatim}

As a final convenience, we also allow integer values to be stored and
incremented or decremented.
Such integer variable updating is achieved by parsing the output string for
occurrences of the pattern {\tt variable\_name += increment} or
{\tt variable\_name -= decrement}
then applying the updates to {\tt variable\_name} in memory.
Importantly, integer addition is also a finite state operation
(see, for example, \cite[Problem 1.32]{sipser13}),
while the Python standard implements the {\tt bignum} type
that handles arbitrarily large integers.
The update function {\tt updates} is shown below.

\begin{verbatim}
def updates(string):
  global MEMORY
  regex = '(\w+)\s*((?:\+|\-)=)\s*(\d+)'
  matches = re.findall(regex, string)
  if matches != None:
    for match in matches:
      label, operator, valuestring = match
      sign = 1 if operator == "+=" else -1
      value = int(valuestring) * sign
      if label in MEMORY and isinstance(MEMORY[label], int):
        MEMORY[label] += value
      else:
        MEMORY[label] = value # new or non-int starts from 0 by default
\end{verbatim}

\subsection{Pre processing language model inputs}

Each input prompt to the language model will be retrieved
from a special memory location {\tt op} and passed as a prompt to
the language model in each computational cycle;
that is, {\tt MEMORY['op']} will serve as an ``instruction register''.
Instruction branching can then be achieved simply
by assigning a different prompt string to {\tt MEMORY['op']}
during a computational cycle.

To access stored memory values, we also allow splicing in the input prompt
string retrieved from {\tt op}.
In particular, the regular expression parser for the input prompt
first detects patterns of the form {\tt @[variable\_name]} and 
replaces these by splicing in the string retrieved from
{\tt MEMORY[variable\_name]}
before passing the prompt string to the language model.
For input pre processing,
it will also be convenient to allow repeated substitutions of nested {\tt @}
occurrences,
so we add the repeated substitution function {\tt substitute\_nested} below.
Note that, technically, allowing arbitrarily nested substitutions
can simulate a context free grammar \cite{sipser13},
which violates the constraint of finite state computation;
however, we will only use bounded depth nesting (depth bound 2)
below to ensure the pre and post processing steps all remain
achievable by finite state computation.

\begin{verbatim}
def substitute_nested(string, char):
  regex = f"(?s)(.*?)(?:{char}\[(\w+)\])(.*)"
  while re.match(regex, string) != None:
    string = substitute(string, char)
  return string
\end{verbatim}

\subsection{Compute cycle}

Finally, a stored instruction computer can run a single compute cycle:
retrieve the next prompt string from {\tt MEMORY['op']};
process the prompt string by possibly splicing in other strings from memory;
pass the prompt string to the language model;
process the output string, possibly splicing in other strings from memory,
detecting all assignments and increment/decrement updates, applying these
to memory;
and repeat.
Computation proceeds until the next instruction in {\tt MEMORY['op']} 
is the special instruction string {\tt 'halt'}.
In particular, the main loop is constructed as follows.

\begin{verbatim}
def main():
  global MEMORY
  while True:
    op = MEMORY['op']
    if op == 'halt':
      return None
    prompt = substitute_nested(op, '@')
    result = call_llm_server(prompt)
    result = substitute(result, '%')
    suffix = assignments(result)
    updates(suffix)
\end{verbatim}

This main instruction loop demonstrates how the language model 
plays the role of the CPU,
effectively
taking the next instruction and its operands expressed by the prompt string
and
acting on these to produce the result string,
which is then used to update the memory.

\section{Universal Turing machine}
\label{sec:turing}

The concept of a universal computer---%
a computing machine
that can simulate the execution of any other computing machine on any input---%
was developed by Alan Turing to solve the \emph{Entscheidungsproblem}
\cite{turing37}.
By the Church-Turing thesis,
all computational mechanisms are considered to be expressible by a
\emph{Turing machine},
which informally consists of a finite state controller
and an unboundedly large ``tape'' memory
with a ``head'' that can access a single tape location
and move one location left or right in each compute cycle
\cite[Chapter 3]{sipser13}.

Formally,
a Turing machine consists of a tuple
${\cal M} = (Q,\Sigma,b,q_0,T,f)$, where
$Q$ is a finite set of states,
$\Sigma$ is a finite set of tape symbols,
$b\in\Sigma$ is the blank symbol,
$q_0\in Q$ is the start state,
$T\subseteq  Q\times\Sigma$ is the set of halting (state, symbol) pairs,
and $f:Q\times\Sigma\rightarrow\Sigma\times\{-1,+1\}\times Q$
is a finite set of transition rules that specify the operation of
the machine in each compute cycle.
We assume the tape is bi-directionally unbounded,
so memory locations can be indexed by an integer $i\in\mathbb{Z}$.
Let $i_0\in\mathbb{Z}$ denote the initial location of the tape head.

The execution of a Turing machine can then be defined as follows.
The tape memory is initialized with a finite number of non-blank symbols
with all other locations blank,
$\cal M$ starts in state $q_0$, and the tape head starts at location $i_0$.
At the start of each compute cycle,
the tape head is at some location $i\in\mathbb{Z}$,
the machine is in some state $q\in Q$,
and some symbol $\sigma\in\Sigma$ is under the tape head.
This combination determines the update
$f(q,\sigma)\mapsto(\sigma', m, q')$,
specifying that
the symbol $\sigma'$ is written at the current memory location $i$,
the machine state $q$ is updated to $q'$,
and the tape head is moved one step left, to location $i'=i-1$ if $m=-1$,
otherwise one step right, to location $i'=i+1$ if $m=+1$.
The compute cycle repeats until the machine encounters a configuration
$(q,\sigma)\in T$.
Non-halting computations are possible.

\cite{shannon56} began
an effort to identify the
smallest universal Turing machines
in terms of the number of states and tape symbols used.
A gap remains between the known upper and lower bounds on
the state and symbol counts for a universal Turing machine
\cite{neary08,nearywoods09},
but progressively smaller universal Turing machines have been identified.
We will consider one such machine in this paper, $U_{15,2}$,
which uses only 15 states and 2 tape symbols \cite{nearywoods09}.
This Turing machine is Pareto optimal in terms of the smallest known
universal Turing machines \cite{neary08}.
Formally, the Turing machine $U_{15,2}$ can be defined by a tuple
$(Q,\Sigma,b,q_0,T,f)$, where
$Q=\{A,B,C,D,E,F,G,H,I,J,K,L,M,N,O\}$,
$\Sigma=\{0,1\}$,
$b=0$,
$q_0=A$,
$T=\{(J,1)\}$,
and the transition function $f$ is defined in Table~\ref{tab:U15,2}.
The initial head position $i_0$ depends
on how the memory is initialized for a given problem instance.

\begin{table}[h]
\centering
\begin{tabular}{c|cccccccc}
    & $A$     & $B$     & $C$     & $D$     & $E$     & $F$     & $G$     & $H$     \\ \hline
$0$ & $0,+,B$ & $1,+,C$ & $0,-,G$ & $0,-,F$ & $1,+,A$ & $1,-,D$ & $0,+,H$ & $1,-,I$ \\
$1$ & $1,+,A$ & $1,+,A$ & $0,-,E$ & $1,-,E$ & $1,-,D$ & $1,-,D$ & $1,-,G$ & $1,-,G$ \\ \hline
\end{tabular}

\begin{tabular}{c|ccccccc}
    & $I$     & $J$     & $K$     & $L$     & $M$     & $N$     & $O$     \\ \hline
$0$ & $0,+,A$ & $1,-,K$ & $0,+,L$ & $0,+,M$ & $0,-,B$ & $0,-,C$ & $0,+,N$ \\
$1$ & $1,-,J$ & halt    & $1,+,N$ & $1,+,L$ & $1,+,L$ & $0,+,O$ & $1,+,N$ \\ \hline
\end{tabular}
\caption{
\em
Transition table for the universal Turing machine $U_{15,2}$.
Rows are indexed by the read symbol $\sigma$,
columns are indexed by the state $q$,
and each table entry $(\sigma',m,q')$ specifies the write symbol $\sigma'$,
the tape head move $m\in\{-1,+1\}$, and the next state $q'$.
}
\label{tab:U15,2}
\end{table}

\section{Simulating $U_{15,2}$ with a prompt program}
\label{sec:program}

In this section, we show that the stored instruction computer
defined in Section~\ref{sec:computer} can be programmed to simulate
the universal Turing machine $U_{15,2}$,
provided that a finite set of conditional assignments and evaluations
can be correctly performed by the language model.
That is, we first propose a specific prompt program that, if executed
correctly, exactly simulates $U_{15,2}$.
The next section will then verify that a specific large language model,
Flan-U-PaLM 540B, is indeed able to execute each of the 
program instructions correctly.

A prompt program consists of a finite set of pre designed strings stored in
memory that provide input prompts to the language model
as part of the main compute cycle,
via a call to {\tt main()}
outlined in Section~\ref{sec:computer}. 
To mimic the behaviour of $U_{15,2}$,
we design the prompt program as follows.
First, a ``boot'' prompt is designed that ``instructs'' the language
model about the behaviour of variable assignments,
variable evaluations after assignment,
and if-then conditionals.

\footnotesize
\begin{verbatim}
boot = """
result = " op="%[B]" %[i]="0" i+=1 "
if 0==1 then result = " op="%[A]" %[i]="1" i+=1 "
$result
" op="%[B]" %[i]="0" i+=1 "

result = " op="%[B]" %[i]="0" i+=1 "
if 1==1 then result = " op="%[A]" %[i]="1" i+=1 "
$result
" op="%[A]" %[i]="1" i+=1 "

result = " op="%[C]" %[i]="1" i+=1 "
if 0==1 then result = " op="%[A]" %[i]="1" i+=1 "
$result
" op="%[C]" %[i]="1" i+=1 "

result = " op="%[C]" %[i]="1" i+=1 "
if 1==1 then result = " op="%[A]" %[i]="1" i+=1 "
$result
" op="%[A]" %[i]="1" i+=1 "

result = " op="%[G]" %[i]="0" i-=1 "
if 0==1 then result = " op="%[E]" %[i]="0" i-=1 "
$result
" op="%[G]" %[i]="0" i-=1 "

result = " op="%[G]" %[i]="0" i-=1 "
if 1==1 then result = " op="%[E]" %[i]="0" i-=1 "
$result
" op="%[E]" %[i]="0" i-=1 "

result = " op="%[F]" %[i]="0" i-=1 "
if 0==1 then result = " op="%[E]" %[i]="1" i-=1 "
$result
" op="%[F]" %[i]="0" i-=1 "

result = " op="%[F]" %[i]="0" i-=1 "
if 1==1 then result = " op="%[E]" %[i]="1" i-=1 "
$result
" op="%[E]" %[i]="1" i-=1 "

result = " op="%[K]" %[i]="1" i-=1 "
if 0==1 then result = " op="halt" "
$result
" op="%[K]" %[i]="1" i-=1 "

result = " op="%[K]" %[i]="1" i-=1 "
if 1==1 then result = " op="halt" "
$result
" op="halt" "

result = " op="%[L]" %[i]="0" i+=1 "
if 0==1 then result = " op="%[N]" %[i]="1" i+=1 "
$result
" op="%[L]" %[i]="0" i+=1 "

result = " op="%[L]" %[i]="0" i+=1 "
if 1==1 then result = " op="%[N]" %[i]="1" i+=1 "
$result
" op="%[N]" %[i]="1" i+=1 "

result = " op="%[C]" %[i]="0" i-=1 "
if 1==1 then result = " op="%[O]" %[i]="0" i+=1 "
$result
" op="%[O]" %[i]="0" i+=1 "

"""
\end{verbatim}
\normalsize

\noindent
Next, a series of ``instruction'' prompts are defined.
Each of these strings is intended to express the logic of a corresponding
Turing machine state in $U_{15,2}$ specified in Table~\ref{tab:U15,2}.

\footnotesize
\begin{verbatim}
A = """@[boot]result = " op="%[B]" %[i]="0" i+=1 "
if @[@[i]]==1 then result = " op="%[A]" %[i]="1" i+=1 "
$result
"""
B = """@[boot]result = " op="%[C]" %[i]="1" i+=1 "
if @[@[i]]==1 then result = " op="%[A]" %[i]="1" i+=1 "
$result
"""
C = """@[boot]result = " op="%[G]" %[i]="0" i-=1 "
if @[@[i]]==1 then result = " op="%[E]" %[i]="0" i-=1 "
$result
"""
D = """@[boot]result = " op="%[F]" %[i]="0" i-=1 "
if @[@[i]]==1 then result = " op="%[E]" %[i]="1" i-=1 "
$result
"""
E = """@[boot]result = " op="%[A]" %[i]="1" i+=1 "
if @[@[i]]==1 then result = " op="%[D]" %[i]="1" i-=1 "
$result
"""
F = """@[boot]result = " op="%[D]" %[i]="1" i-=1 "
$result
"""
G = """@[boot]result = " op="%[H]" %[i]="0" i-=1 "
if @[@[i]]==1 then result = " op="%[G]" %[i]="1" i-=1 "
$result
"""
H = """@[boot]result = " op="%[I]" %[i]="1" i-=1 "
if @[@[i]]==1 then result = " op="%[G]" %[i]="1" i-=1 "
$result
"""
I = """@[boot]result = " op="%[A]" %[i]="0" i+=1 "
if @[@[i]]==1 then result = " op="%[J]" %[i]="1" i-=1 "
$result
"""
J = """@[boot]result = " op="%[K]" %[i]="1" i-=1 "
if @[@[i]]==1 then result = " op="halt" "
$result
"""
K = """@[boot]result = " op="%[L]" %[i]="0" i+=1 "
if @[@[i]]==1 then result = " op="%[N]" %[i]="1" i+=1 "
$result
"""
L = """@[boot]result = " op="%[M]" %[i]="0" i+=1 "
if @[@[i]]==1 then result = " op="%[L]" %[i]="1" i+=1 "
$result
"""
M = """@[boot]result = " op="%[B]" %[i]="0" i-=1 "
if @[@[i]]==1 then result = " op="%[L]" %[i]="1" i+=1 "
$result
"""
N = """@[boot]result = " op="%[C]" %[i]="0" i-=1 "
if @[@[i]]==1 then result = " op="%[O]" %[i]="0" i+=1 "
$result
"""
O = """@[boot]result = " op="%[N]" %[i]="0" i+=1 "
if @[@[i]]==1 then result = " op="%[N]" %[i]="1" i+=1 "
$result
"""
\end{verbatim}
\normalsize

It helps to understand how this prompt program is intended to work.
First, note that the memory location {\tt 'i'} is intended to keep track of
the current location of the Turing machine head,
so that any update {\tt i-=1} will correspond to moving the head one step
left, and {\tt i+=1} will correspond to moving the head one step right.
Next, consider the post processing of one of the result strings,
for example {\tt " op=\%[N]" \%[i]="1" i+=1 "}.
In this string,
the expression {\tt \%[i]="1"} is intended to write the target symbol
{\tt '1'} to the memory location indexed by {\tt MEMORY['i']}.
That is, during post processing, any substring of the form {\tt \%[x]}
will first be replaced by the string in {\tt MEMORY['x']}
before performing any assignments, as explained in Section~\ref{sec:computer}.
Thus, given {\tt \%[i]="1"} the substring {\tt \%[i]} will first be replaced by
the value in {\tt MEMORY['i']},
say $\ell$,
which then serves as the label in memory to be assigned the value {\tt '1'}.
For example,
if we assume {\tt MEMORY['i'] = 42} and {\tt MEMORY['42'] = '0'},
then after post processing and assignment we will have {\tt MEMORY['42'] = '1'}.
Control branching, i.e., a state transition, is achieved by assigning 
a new instruction string from \{{\tt A}, ..., {\tt O}\}
to the instruction register {\tt MEMORY['op']},
as specified by the assignment string {\tt op="\%[N]} in the example.

In the pre processing phase, the prompt string is obtained
from {\tt MEMORY['op']},
then substrings of the form {\tt @[x]} are 
replaced by the string stored in {\tt MEMORY['x']}.
This allows for more compact instruction strings
{\tt A}, ..., {\tt O},
since the lengthy {\tt boot} string can just be spliced in during
pre processing.
More importantly,
the symbol at the current head position can be read with {\tt @[@[i]]}.
To see why this works,
note that the preprocessor will apply \emph{nested} substitutions of the
{\tt @[x]} patterns, as discussed in Section~\ref{sec:computer}.
Therefore, if we continue to assume that {\tt MEMORY['i'] = 42} and
{\tt MEMORY['42'] = '1'},
the first substitution of {\tt @[@[i]]} will result in {\tt @['42']},
and the second substitution will result in {\tt '1'}.
That is, after pre processing, the substring {\tt @[@[i]]} is replaced
with the value found in {\tt MEMORY[MEMORY['i']]}, 
i.e., the symbol at the current position of the tape head,
which in this case will be {\tt '1'}.

Given this understanding, it is easy to verify that each of the
instruction strings {\tt A}, ..., {\tt O} 
correctly mimics the logic of the corresponding states
$A$, ..., $O$ in Table~\ref{tab:U15,2},
including conditioning on the current symbol in the head position,
writing the correct symbol to the current head position,
moving the head position in the correct direction,
and updating the state by assigning the correct next instruction to
{\tt 'op'}.

Finally, to simulate the behaviour of $U_{15,2}$,
we also have to consider the initial contents of the tape memory.
Let the variable {\tt TAPE} be assigned to a string 
that covers the non-blank portion of the initial memory for $U_{15,2}$,
which must be finitely long by definition.
Also let {\tt i}$=i_0$ contain the initial position of the tape head.
To simulate the Turing machine from this configuration,
we first initialize the associative memory, {\tt MEMORY}, as follows,
then simply call {\tt main()} as specified in Section~\ref{sec:computer}.

\newpage

\begin{verbatim}
MEMORY = {'boot':boot}
for s in 'ABCDEFGHIJKLMNO':
  MEMORY[s] = eval(s)
for loc in range(len(TAPE)):
  MEMORY[str(loc)] = TAPE[loc]
BLANK = '0'
MEMORY['i'] = i
MEMORY['op'] = A
main()
\end{verbatim}

We now claim that each compute cycle of {\tt main()}
maintains an exact equivalence with each compute cycle of 
the Turing machine $U_{15,2}$.
The proof is by induction on the number of compute cycles.

At initialization, there is an equivalence between:
the contents of the Turing machine memory tape and the 
contents of {\tt MEMORY} labelled by location numbers
(with all unassigned locations assumed to be blank);
the initial tape head location $i_0$ and {\tt i};
the initial state $A$ and initial instruction {\tt A}.

Then, inductively, assume the same equivalence holds at the onset
of a compute cycle.
The Turing machine will then be updated according to
$(q,\sigma)\mapsto(\sigma',m,q')$
following the specification in Table~\ref{tab:U15,2}.
Now assume the corresponding prompt string
after pre processing with the same current symbol in {\tt MEMORY[MEMORY['i']]}
(i.e., the current location of the simulated tape head)
returns the correct result string
after the call to {\tt call\_llm\_server(prompt)}.
Then one can verify, on a case by case basis for each (state, symbol) pair, 
that the result string
specifies the same symbol to be written to the current head location,
specifies the same direction to move the head,
and specifies the corresponding next instruction,
thus the equivalence is maintained at the end of the cycle.

To illustrate one of the (state, symbol) verifications,
consider the first entry in Table~\ref{tab:U15,2},
which specifies the Turing machine update $(A,0)\mapsto(0,+,B)$.
Observe that the instruction {\tt A} pre processed with the same
input symbol {\tt '0'} at {\tt MEMORY[MEMORY['i']]}
will return the result string {\tt " op=\%[B]" \%[i]="0" i+=1 "}
assuming the language model operates correctly (verified below).
In this case, the post processing phase will
write the corresponding symbol {\tt '0'} to the current memory location,
move the head right {\tt +=1}, and assign {\tt B} to be the next instruction,
thus maintaining the equivalence.

Similarly, if the current input is $1$, the Turing machine update is
$(A,1)\mapsto(1,+,A)$.
In this case,
observe that if the instruction {\tt A} pre processed with the same
input symbol {\tt '1'} at {\tt MEMORY[MEMORY['i']]},
the condition will be true and the result string will be
{\tt " op=\%[A]" \%[i]="1" i+=1 "}
assuming the language model operates correctly.
Post processing will once again maintain the equivalence.

A similar verification succeeds for all 29 (state, symbol) cases.
(Note that the update in state $F$ does not depend on the input,
so there is one fewer case than the total number of (state, symbol) pairs.)

It remains only to verify that a language model 
can produce the correct result string
given any of the instruction strings after pre processing
with the current memory symbol.

\section{Verifying correct execution using Flan-U-PaLM 540B}
\label{sec:verify}

We now consider the specific language model
Flan-U-PaLM 540B \cite{chungetal22},
which is a large 540B parameter model
that has been refined with additional instruction fine-tuning.
To ensure deterministic computation,
the decoding temperature for the language model
is set to zero (pure greedy decoding).

To complete the argument,
we now simply enumerate each of the possible (state, symbol) combinations
and verify that the language model produces the correct result string
given an input prompt string composed from the corresponding instruction
and pre processed with the corresponding input symbol.
This is simply a brute force proof, calling the language model with
each possible input prompt and verifying that the correct result string
is indeed returned.
There are 29 cases. 
(So, yeah, human readable but not human enjoyable, apologies.)

\bigskip

\noindent
Verification test 1 (state A read 0)
\scriptsize
\begin{verbatim}
head = MEMORY['i']
MEMORY[str(head)] = '0'
op = A
prompt = substitute_nested(op, '@')
result = call_llm_server(prompt)
print(op)
print(result)
\end{verbatim}
\normalsize

\noindent
Output 1
\scriptsize
\begin{verbatim}
@[boot]result = " op="%[B]" %[i]="0" i+=1 "
if @[@[i]]==1 then result = " op="%[A]" %[i]="1" i+=1 "
$result

" op="%[B]" %[i]="0" i+=1 "
\end{verbatim}
\normalsize

\noindent
Verification test 2 (state A read 1)
\scriptsize
\begin{verbatim}
head = MEMORY['i']
MEMORY[str(head)] = '1'
op = A
prompt = substitute_nested(op, '@')
result = call_llm_server(prompt)
print(op)
print(result)
\end{verbatim}
\normalsize

\noindent
Output 2
\scriptsize
\begin{verbatim}
@[boot]result = " op="%[B]" %[i]="0" i+=1 "
if @[@[i]]==1 then result = " op="%[A]" %[i]="1" i+=1 "
$result

" op="%[A]" %[i]="1" i+=1 "
\end{verbatim}
\normalsize

\noindent
Verification test 3 (state B read 0)
\scriptsize
\begin{verbatim}
head = MEMORY['i']
MEMORY[str(head)] = '0'
op = B
prompt = substitute_nested(op, '@')
result = call_llm_server(prompt)
print(op)
print(result)
\end{verbatim}
\normalsize

\noindent
Output 3
\scriptsize
\begin{verbatim}
@[boot]result = " op="%[C]" %[i]="1" i+=1 "
if @[@[i]]==1 then result = " op="%[A]" %[i]="1" i+=1 "
$result

" op="%[C]" %[i]="1" i+=1 "
\end{verbatim}
\normalsize

\noindent
Verification test 4 (state B read 1)
\scriptsize
\begin{verbatim}
head = MEMORY['i']
MEMORY[str(head)] = '1'
op = B
prompt = substitute_nested(op, '@')
result = call_llm_server(prompt)
print(op)
print(result)
\end{verbatim}
\normalsize

\noindent
Output 4
\scriptsize
\begin{verbatim}
@[boot]result = " op="%[C]" %[i]="1" i+=1 "
if @[@[i]]==1 then result = " op="%[A]" %[i]="1" i+=1 "
$result

" op="%[A]" %[i]="1" i+=1 "
\end{verbatim}
\normalsize

\noindent
Verification test 5 (state C read 0)
\scriptsize
\begin{verbatim}
head = MEMORY['i']
MEMORY[str(head)] = '0'
op = C
prompt = substitute_nested(op, '@')
result = call_llm_server(prompt)
print(op)
print(result)
\end{verbatim}
\normalsize

\noindent
Output 5
\scriptsize
\begin{verbatim}
@[boot]result = " op="%[G]" %[i]="0" i-=1 "
if @[@[i]]==1 then result = " op="%[E]" %[i]="0" i-=1 "
$result

" op="%[G]" %[i]="0" i-=1 "
\end{verbatim}
\normalsize

\noindent
Verification test 6 (state C read 1)
\scriptsize
\begin{verbatim}
head = MEMORY['i']
MEMORY[str(head)] = '1'
op = C
prompt = substitute_nested(op, '@')
result = call_llm_server(prompt)
print(op)
print(result)
\end{verbatim}
\normalsize

\noindent
Output 6
\scriptsize
\begin{verbatim}
@[boot]result = " op="%[G]" %[i]="0" i-=1 "
if @[@[i]]==1 then result = " op="%[E]" %[i]="0" i-=1 "
$result

" op="%[E]" %[i]="0" i-=1 "
\end{verbatim}
\normalsize

\noindent
Verification test 7 (state D read 0)
\scriptsize
\begin{verbatim}
head = MEMORY['i']
MEMORY[str(head)] = '0'
op = D
prompt = substitute_nested(op, '@')
result = call_llm_server(prompt)
print(op)
print(result)
\end{verbatim}
\normalsize

\noindent
Output 7
\scriptsize
\begin{verbatim}
@[boot]result = " op="%[F]" %[i]="0" i-=1 "
if @[@[i]]==1 then result = " op="%[E]" %[i]="1" i-=1 "
$result

" op="%[F]" %[i]="0" i-=1 "
\end{verbatim}
\normalsize

\noindent
Verification test 8 (state D read 1)
\scriptsize
\begin{verbatim}
head = MEMORY['i']
MEMORY[str(head)] = '1'
op = D
prompt = substitute_nested(op, '@')
result = call_llm_server(prompt)
print(op)
print(result)
\end{verbatim}
\normalsize

\noindent
Output 8
\scriptsize
\begin{verbatim}
@[boot]result = " op="%[F]" %[i]="0" i-=1 "
if @[@[i]]==1 then result = " op="%[E]" %[i]="1" i-=1 "
$result

" op="%[E]" %[i]="1" i-=1 "
\end{verbatim}
\normalsize

\noindent
Verification test 9 (state E read 0)
\scriptsize
\begin{verbatim}
head = MEMORY['i']
MEMORY[str(head)] = '0'
op = E
prompt = substitute_nested(op, '@')
result = call_llm_server(prompt)
print(op)
print(result)
\end{verbatim}
\normalsize

\noindent
Output 9
\scriptsize
\begin{verbatim}
@[boot]result = " op="%[A]" %[i]="1" i+=1 "
if @[@[i]]==1 then result = " op="%[D]" %[i]="1" i-=1 "
$result

" op="%[A]" %[i]="1" i+=1 "
\end{verbatim}
\normalsize

\noindent
Verification test 10 (state E read 1)
\scriptsize
\begin{verbatim}
head = MEMORY['i']
MEMORY[str(head)] = '1'
op = E
prompt = substitute_nested(op, '@')
result = call_llm_server(prompt)
print(op)
print(result)
\end{verbatim}
\normalsize

\noindent
Output 10
\scriptsize
\begin{verbatim}
@[boot]result = " op="%[A]" %[i]="1" i+=1 "
if @[@[i]]==1 then result = " op="%[D]" %[i]="1" i-=1 "
$result

" op="%[D]" %[i]="1" i-=1 "
\end{verbatim}
\normalsize

\noindent
Verification test 11 (state F read 0 == state F read 1)
\scriptsize
\begin{verbatim}
op = F
prompt = substitute_nested(op, '@')
result = call_llm_server(prompt)
print(op)
print(result)
\end{verbatim}
\normalsize

\noindent
Output 11
\scriptsize
\begin{verbatim}
@[boot]result = " op="%[D]" %[i]="1" i-=1 "
$result

op="%[D]" %[i]="1" i-=1
\end{verbatim}
\normalsize

\noindent
Verification test 12 (state G read 0)
\scriptsize
\begin{verbatim}
head = MEMORY['i']
MEMORY[str(head)] = '0'
op = G
prompt = substitute_nested(op, '@')
result = call_llm_server(prompt)
print(op)
print(result)
\end{verbatim}
\normalsize

\noindent
Output 12
\scriptsize
\begin{verbatim}
@[boot]result = " op="%[H]" %[i]="0" i-=1 "
if @[@[i]]==1 then result = " op="%[G]" %[i]="1" i-=1 "
$result

" op="%[H]" %[i]="0" i-=1 "
\end{verbatim}
\normalsize

\noindent
Verification test 13 (state G read 1)
\scriptsize
\begin{verbatim}
head = MEMORY['i']
MEMORY[str(head)] = '1'
op = G
prompt = substitute_nested(op, '@')
result = call_llm_server(prompt)
print(op)
print(result)
\end{verbatim}
\normalsize

\noindent
Output 13
\scriptsize
\begin{verbatim}
@[boot]result = " op="%[H]" %[i]="0" i-=1 "
if @[@[i]]==1 then result = " op="%[G]" %[i]="1" i-=1 "
$result

" op="%[G]" %[i]="1" i-=1 "
\end{verbatim}
\normalsize

\noindent
Verification test 14 (state H read 0)
\scriptsize
\begin{verbatim}
head = MEMORY['i']
MEMORY[str(head)] = '0'
op = H
prompt = substitute_nested(op, '@')
result = call_llm_server(prompt)
print(op)
print(result)
\end{verbatim}
\normalsize

\noindent
Output 14
\scriptsize
\begin{verbatim}
@[boot]result = " op="%[I]" %[i]="1" i-=1 "
if @[@[i]]==1 then result = " op="%[G]" %[i]="1" i-=1 "
$result

" op="%[I]" %[i]="1" i-=1 "
\end{verbatim}
\normalsize

\noindent
Verification test 15 (state H read 1)
\scriptsize
\begin{verbatim}
head = MEMORY['i']
MEMORY[str(head)] = '1'
op = H
prompt = substitute_nested(op, '@')
result = call_llm_server(prompt)
print(op)
print(result)
\end{verbatim}
\normalsize

\noindent
Output 15
\scriptsize
\begin{verbatim}
@[boot]result = " op="%[I]" %[i]="1" i-=1 "
if @[@[i]]==1 then result = " op="%[G]" %[i]="1" i-=1 "
$result

" op="%[G]" %[i]="1" i-=1 "
\end{verbatim}
\normalsize

\noindent
Verification test 16 (state I read 0)
\scriptsize
\begin{verbatim}
head = MEMORY['i']
MEMORY[str(head)] = '0'
op = I
prompt = substitute_nested(op, '@')
result = call_llm_server(prompt)
print(op)
print(result)
\end{verbatim}
\normalsize

\noindent
Output 16
\scriptsize
\begin{verbatim}
@[boot]result = " op="%[A]" %[i]="0" i+=1 "
if @[@[i]]==1 then result = " op="%[J]" %[i]="1" i-=1 "
$result

" op="%[A]" %[i]="0" i+=1 "
\end{verbatim}
\normalsize

\noindent
Verification test 17 (state I read 1)
\scriptsize
\begin{verbatim}
head = MEMORY['i']
MEMORY[str(head)] = '1'
op = I
prompt = substitute_nested(op, '@')
result = call_llm_server(prompt)
print(op)
print(result)
\end{verbatim}
\normalsize

\noindent
Output 17
\scriptsize
\begin{verbatim}
@[boot]result = " op="%[A]" %[i]="0" i+=1 "
if @[@[i]]==1 then result = " op="%[J]" %[i]="1" i-=1 "
$result

" op="%[J]" %[i]="1" i-=1 "
\end{verbatim}
\normalsize

\noindent
Verification test 18 (state J read 0)
\scriptsize
\begin{verbatim}
head = MEMORY['i']
MEMORY[str(head)] = '0'
op = J
prompt = substitute_nested(op, '@')
result = call_llm_server(prompt)
print(op)
print(result)
\end{verbatim}
\normalsize

\noindent
Output 18
\scriptsize
\begin{verbatim}
@[boot]result = " op="%[K]" %[i]="1" i-=1 "
if @[@[i]]==1 then result = " op="halt" "
$result

" op="%[K]" %[i]="1" i-=1 "
\end{verbatim}
\normalsize

\noindent
Verification test 19 (state J read 1)
\scriptsize
\begin{verbatim}
head = MEMORY['i']
MEMORY[str(head)] = '1'
op = J
prompt = substitute_nested(op, '@')
result = call_llm_server(prompt)
print(op)
print(result)
\end{verbatim}
\normalsize

\noindent
Output 19
\scriptsize
\begin{verbatim}
@[boot]result = " op="%[K]" %[i]="1" i-=1 "
if @[@[i]]==1 then result = " op="halt" "
$result

" op="halt" "
\end{verbatim}
\normalsize

\noindent
Verification test 20 (state K read 0)
\scriptsize
\begin{verbatim}
head = MEMORY['i']
MEMORY[str(head)] = '0'
op = K
prompt = substitute_nested(op, '@')
result = call_llm_server(prompt)
print(op)
print(result)
\end{verbatim}
\normalsize

\noindent
Output 20
\scriptsize
\begin{verbatim}
@[boot]result = " op="%[L]" %[i]="0" i+=1 "
if @[@[i]]==1 then result = " op="%[N]" %[i]="1" i+=1 "
$result

" op="%[L]" %[i]="0" i+=1 "
\end{verbatim}
\normalsize

\noindent
Verification test 21 (state K read 1)
\scriptsize
\begin{verbatim}
head = MEMORY['i']
MEMORY[str(head)] = '1'
op = K
prompt = substitute_nested(op, '@')
result = call_llm_server(prompt)
print(op)
print(result)
\end{verbatim}
\normalsize

\noindent
Output 21
\scriptsize
\begin{verbatim}
@[boot]result = " op="%[L]" %[i]="0" i+=1 "
if @[@[i]]==1 then result = " op="%[N]" %[i]="1" i+=1 "
$result

" op="%[N]" %[i]="1" i+=1 "
\end{verbatim}
\normalsize

\noindent
Verification test 22 (state L read 0)
\scriptsize
\begin{verbatim}
head = MEMORY['i']
MEMORY[str(head)] = '0'
op = L
prompt = substitute_nested(op, '@')
result = call_llm_server(prompt)
print(op)
print(result)
\end{verbatim}
\normalsize

\noindent
Output 22
\scriptsize
\begin{verbatim}
@[boot]result = " op="%[M]" %[i]="0" i+=1 "
if @[@[i]]==1 then result = " op="%[L]" %[i]="1" i+=1 "
$result

" op="%[M]" %[i]="0" i+=1 "
\end{verbatim}
\normalsize

\noindent
Verification test 23 (state L read 1)
\scriptsize
\begin{verbatim}
head = MEMORY['i']
MEMORY[str(head)] = '1'
op = L
prompt = substitute_nested(op, '@')
result = call_llm_server(prompt)
print(op)
print(result)
\end{verbatim}
\normalsize

\noindent
Output 23
\scriptsize
\begin{verbatim}
@[boot]result = " op="%[M]" %[i]="0" i+=1 "
if @[@[i]]==1 then result = " op="%[L]" %[i]="1" i+=1 "
$result

" op="%[L]" %[i]="1" i+=1 "
\end{verbatim}
\normalsize

\noindent
Verification test 24 (state M read 0)
\scriptsize
\begin{verbatim}
head = MEMORY['i']
MEMORY[str(head)] = '0'
op = M
prompt = substitute_nested(op, '@')
result = call_llm_server(prompt)
print(op)
print(result)
\end{verbatim}
\normalsize

\noindent
Output 24
\scriptsize
\begin{verbatim}
@[boot]result = " op="%[B]" %[i]="0" i-=1 "
if @[@[i]]==1 then result = " op="%[L]" %[i]="1" i+=1 "
$result

" op="%[B]" %[i]="0" i-=1 "
\end{verbatim}
\normalsize

\noindent
Verification test 25 (state M read 1)
\scriptsize
\begin{verbatim}
head = MEMORY['i']
MEMORY[str(head)] = '1'
op = M
prompt = substitute_nested(op, '@')
result = call_llm_server(prompt)
print(op)
print(result)
\end{verbatim}
\normalsize

\noindent
Output 25
\scriptsize
\begin{verbatim}
@[boot]result = " op="%[B]" %[i]="0" i-=1 "
if @[@[i]]==1 then result = " op="%[L]" %[i]="1" i+=1 "
$result

" op="%[L]" %[i]="1" i+=1 "
\end{verbatim}
\normalsize

\noindent
Verification test 26 (state N read 0)
\scriptsize
\begin{verbatim}
head = MEMORY['i']
MEMORY[str(head)] = '0'
op = N
prompt = substitute_nested(op, '@')
result = call_llm_server(prompt)
print(op)
print(result)
\end{verbatim}
\normalsize

\noindent
Output 26
\scriptsize
\begin{verbatim}
@[boot]result = " op="%[C]" %[i]="0" i-=1 "
if @[@[i]]==1 then result = " op="%[O]" %[i]="0" i+=1 "
$result

" op="%[C]" %[i]="0" i-=1 "
\end{verbatim}
\normalsize

\noindent
Verification test 27 (state N read 1)
\scriptsize
\begin{verbatim}
head = MEMORY['i']
MEMORY[str(head)] = '1'
op = N
prompt = substitute_nested(op, '@')
result = call_llm_server(prompt)
print(op)
print(result)
\end{verbatim}
\normalsize

\noindent
Output 27
\scriptsize
\begin{verbatim}
@[boot]result = " op="%[C]" %[i]="0" i-=1 "
if @[@[i]]==1 then result = " op="%[O]" %[i]="0" i+=1 "
$result

" op="%[O]" %[i]="0" i+=1 "
\end{verbatim}
\normalsize

\noindent
Verification test 28 (state O read 0)
\scriptsize
\begin{verbatim}
head = MEMORY['i']
MEMORY[str(head)] = '0'
op = O
prompt = substitute_nested(op, '@')
result = call_llm_server(prompt)
print(op)
print(result)
\end{verbatim}
\normalsize

\noindent
Output 28
\scriptsize
\begin{verbatim}
@[boot]result = " op="%[N]" %[i]="0" i+=1 "
if @[@[i]]==1 then result = " op="%[N]" %[i]="1" i+=1 "
$result

" op="%[N]" %[i]="0" i+=1 "
\end{verbatim}
\normalsize

\noindent
Verification test 29 (state O read 1)
\scriptsize
\begin{verbatim}
head = MEMORY['i']
MEMORY[str(head)] = '1'
op = O
prompt = substitute_nested(op, '@')
result = call_llm_server(prompt)
print(op)
print(result)
\end{verbatim}
\normalsize

\noindent
Output 29
\scriptsize
\begin{verbatim}
@[boot]result = " op="%[N]" %[i]="0" i+=1 "
if @[@[i]]==1 then result = " op="%[N]" %[i]="1" i+=1 "
$result

" op="%[N]" %[i]="1" i+=1 "
\end{verbatim}
\normalsize

\noindent
This completes the proof.

\section{Discussion}

Hopefully the reader has been convinced by this point.
There are some reflections on this study that are useful to share.

Although the verification of computational universality is straightforward,
the language model's behaviour was brittle.
Success was not achieved with every large language model considered,
and effort was required to engineer the prompts.
For example, the instruction strings {\tt A}, ..., {\tt O}
with the substitution symbols {\tt @} and {\tt \%} 
are terse and not particularly elegant for humans to read,
yet compactness seemed essential
to get the language model to produce correct results.
Eliciting correct evaluations of variable assignments
was sometimes a challenge, but by far the biggest challenge was
getting the language model to interpret conditionals properly.
The reader might notice that the conditionals have been reduced to
if-then rather than if-then-else forms.
This was not an accident:
I was not able to get the language model to reliably produce
correct outputs for if-then-else conditionals.
The difficulty with conditionals also makes it challenging to 
simulate other, smaller universal Turing machines,
such as $U_{6,4}$ \cite{nearywoods09},
since this requires a series of 3 conditionals for each state,
which I could not get to work without introducing phantom states
and reducing instructions to only a single conditional,
ultimately ending up with a far less digestible construction.
Presumably improvements in the underlying language models will
mitigate such challenges.

Earlier versions of this work considered simulating Rule 110 for
a one dimensional cellular automaton \cite{wolfram02},
leveraging the fact that this is known to be a (weakly) Turing
complete \cite{cook04}.
Although far more visually appealing, Rule 110 requires an unbounded
periodic initialization to an simulate arbitrary Turing machine,
and ultimately the more direct simulation of $U_{15,2}$ presented
in this paper, which requires only a bounded initialization,
appears to be more convincing.

There is an interesting analogy to the ``programming language''
developed in Sections~\ref{sec:computer} and \ref{sec:program}
and some of the earliest programming languages \cite{boehm54},
including the first assembly languages \cite{boothbritten47}.
The latter is particularly reminiscent given the reliance on human readable
labels for branch control.
It is interesting to speculate about what other concepts in the
history of software engineering
(e.g., high level languages, modularity, libraries, etc.)
might be useful for eliciting desired computational behaviour
from a large language model.

The result in this paper is distinct from previous studies that investigate
the computational universality of neural sequence models,
such as recurrent neural networks
\\
\cite{siegelmannsontag92,weissetal18}
and Transformers
\cite{perezetal19},
\\
\cite{bhattamishraetal20,weietal22a}.
The key distinction is that we consider a \emph{fixed} language model
with frozen weights, 
and show how external memory augmentation can elicit universal
computational behaviour.
By contrast these past studies have shown how
computationally universal behaviour can be recovered by
\emph{manipulating} the weights of the neural network,
typically using unbounded (or sufficiently high) precision
weights to encode data structures,
like multiple stacks.
An advantage of these past works is that they do not require any external
memory to demonstrate universal computational behaviour.
On the other hand, these results do not apply to existing
large language models without altering their weights
(as far as currently known).
The results in this paper show that large language models are already
computationally universal---as they exist currently---provided only
that they have access to an unbounded external memory.

\section*{Acknowledgments}

Sincere thanks to
Noah~Fiedel,
Ramki Gummadi,
Andr\'{a}s~Gy\"{o}rgy,
Chris~Harris,
Tengyu~Ma,
Jason~Wei,
Sherry~Yang,
Denny~Zhou
and
Martin~Zinkevich
for essential discussions leading to this work.
Thanks also to Google Brain and my team members for providing
an ideal environment for conducting exploratory research.
Support from the CIFAR Canada AI Research Chairs program, NSERC, and Amii
is also gratefully acknowledged.

\bibliographystyle{apalike}

\end{document}